\documentclass{article}

\usepackage{microtype}
\usepackage{graphicx}
\usepackage{subcaption}
\usepackage{booktabs} 

\usepackage{hyperref}

\usepackage[accepted]{icml2024}

\usepackage{amsmath,amsfonts,bm}

\def\eqref#1{equation~\ref{#1}}

\def\1{\bm{1}}

\def\rb{{\textnormal{b}}}

\def\rva{{\mathbf{a}}}

\def\rvg{{\mathbf{g}}}

\def\rvs{{\mathbf{s}}}

\def\rvx{{\mathbf{x}}}

\def\vb{{\bm{b}}}

\DeclareMathAlphabet{\mathsfit}{\encodingdefault}{\sfdefault}{m}{sl}
\SetMathAlphabet{\mathsfit}{bold}{\encodingdefault}{\sfdefault}{bx}{n}

\newcommand{\E}{\mathbb{E}}
\newcommand{\Ls}{\mathcal{L}}

\let\ab\allowbreak

\usepackage[utf8]{inputenc} 
\usepackage[T1]{fontenc}    
\usepackage{url}            
\usepackage{nicefrac}       
\usepackage{xcolor}         
\usepackage{lipsum}
\usepackage{adjustbox}
\usepackage{algorithm}

\usepackage{pgfplots}
\usepackage{graphicx}
\usepackage[font=small,labelfont=bf]{caption}
\usepackage{multirow}
\usepackage{pifont}

\definecolor{amethyst}{rgb}{0.6, 0.4, 0.8}
\definecolor{atomictangerine}{rgb}{1.0, 0.6, 0.4}
\definecolor{ballblue}{rgb}{0.13, 0.67, 0.8}
\definecolor{coral}{rgb}{1.0, 0.5, 0.31}
\definecolor{carnationpink}{rgb}{1.0, 0.65, 0.79}
\definecolor{lightcoral}{rgb}{0.94, 0.5, 0.5}
\definecolor{darkcyan}{rgb}{0.0, 0.55, 0.55}
\definecolor{darkcoral}{rgb}{0.8, 0.36, 0.27}
\definecolor{green(ryb)}{rgb}{0.4, 0.69, 0.2}
\definecolor{darktangerine}{rgb}{1.0, 0.66, 0.07}
\definecolor{fluorescentorange}{rgb}{1.0, 0.75, 0.0}
\definecolor{bluegray}{rgb}{0.5, 0.55, 0.64}
\definecolor{xanthous}{rgb}{1, 0.6, 0.2}

\newcommand*\colourcheck[1]{%
  \expandafter\newcommand\csname #1check\endcsname{\textcolor{#1}{\ding{52}}}%
}
\colourcheck{green}

\newcommand*\colourxmark[1]{%
  \expandafter\newcommand\csname #1xmark\endcsname{\textcolor{#1}{\ding{53}}}%
}
\colourxmark{red}

\usepackage{amsmath}
\usepackage{amssymb}
\usepackage{mathtools}
\usepackage{amsthm}

\def\rva{{\mathbf{a}}}

\def\rvg{{\mathbf{g}}}

\def\rvs{{\mathbf{s}}}

\def\rvx{{\mathbf{x}}}

\usepackage[capitalize,noabbrev]{cleveref}

\theoremstyle{plain}

\theoremstyle{definition}

\newtheorem{example}{Example}
\theoremstyle{remark}

\begin{document}

\twocolumn[
\icmltitle{PlanDQ: Hierarchical Plan Orchestration via D-Conductor and Q-Performer}



\icmlsetsymbol{equal}{*}

\begin{icmlauthorlist}
\icmlauthor{Chang Chen}{rutgers}
\icmlauthor{Junyeob Baek}{kaist}
\icmlauthor{Fei Deng}{rutgers}
\icmlauthor{Kenji Kawaguchi}{nus}
\icmlauthor{Caglar Gulcehre}{epfl}
\icmlauthor{Sungjin Ahn}{kaist}
\end{icmlauthorlist}

\icmlaffiliation{rutgers}{Rutgers University}
\icmlaffiliation{kaist}{KAIST}
\icmlaffiliation{epfl}{EPFL}
\icmlaffiliation{nus}{National University of Singapore}

\icmlcorrespondingauthor{Chang Chen}{chang.chen@rutgers.edu}
\icmlcorrespondingauthor{Sungjin Ahn}{sungjin.ahn@kaist.ac.kr}

\icmlkeywords{Machine Learning, ICML}

\vskip 0.3in
]



\printAffiliationsAndNotice{} 

\begin{abstract}
Despite the recent advancements in offline RL, no unified algorithm could achieve superior performance across a broad range of tasks. Offline \textit{value function learning}, in particular, struggles with sparse-reward, long-horizon tasks due to the difficulty of solving credit assignment and extrapolation errors that accumulates as the horizon of the task grows.~On the other hand, models that can perform well in long-horizon tasks are designed specifically for goal-conditioned tasks, which commonly perform worse than value function learning methods on short-horizon, dense-reward scenarios. To bridge this gap, we propose a hierarchical planner designed for offline RL called PlanDQ. PlanDQ incorporates a diffusion-based planner at the high level, named D-Conductor, which guides the low-level policy through sub-goals. At the low level, we used a Q-learning based approach called the Q-Performer to accomplish these sub-goals. Our experimental results suggest that PlanDQ can achieve superior or competitive performance on D4RL continuous control benchmark tasks as well as AntMaze, Kitchen, and Calvin as long-horizon tasks.
\end{abstract}

\section{Introduction}
\label{main:intro}

Offline reinforcement learning (RL) \citep{agarwal2020optimistic,gulcehre2020rl} aims to address the challenges of learning policies from a fixed offline dataset without needing additional costly online interactions with the environment. This ability is particularly important in scenarios where real-time interaction is either impractical or risky, such as healthcare, autonomous driving, and robotics \citep{sinha2021srl, de2021discovering, tang2022leveraging}. However, offline RL presents unique challenges, the most significant of which is the distributional shift between the behavior policy and the learned policy. Additionally, the quality and diversity of the offline data significantly impact the performance and generalizability of the learned policies \citep{levine2020offline, bhargava2023sequence}.

In response to these challenges, recent advances in policy learning highlighted the benefits of integrating expressive diffusion models into offline RL. Among those works, the pioneering works like Diffuser \citep{janner2022diffuser} and subsequent models \citep{hdmi, hong2023diffused, chen2024simple} focused on modeling trajectory distributions. Thus, planning with these learned models is equivalent to sampling from them, showing notable improvements, especially in domains requiring long-horizon planning. However, these diffusion planners often fall short in tasks characterized by short horizons and dense rewards.  In contrast, another line of research, utilizing diffusion models to represent complex behavior policies \citep{wang2023diffusion, cheng2023cep}, are optimized for maximizing expected Q values and show superior performance in dense reward settings. Nonetheless, these value-function learning methods often exhibit limitations in long-horizon, sparse-reward scenarios due to the distinct learning paradigms they employ.

As summarized in Table \ref{table:intro_analysis}, value-based methods update the value function at each timestep independently, propagating optimal future rewards backward. Although effective in short-horizon, dense-reward settings, the value propagation over lengthy time horizons poses inherent challenges \citep{bhargava2023sequence}, necessitating the hierarchical structure in long-horizon tasks \citep{park2023hiql, hdmi, hong2023diffused, chen2024simple}. On the other hand, sequence modeling methods struggle with stitching optimal policies from sub-optimal trajectories in short-horizon tasks with dense rewards, an essential skill for offline RL \citep{yamagata2023q, kumar2022should, d4rl, hong2023offline}. This poses difficulty in learning an efficient low-level policy due to the short-horizon property of sub-tasks. Moreover, our empirical observations indicate that, without Bellman updating, the guidance function in diffusion-based sequence modeling methods tends to converge to sub-optimal solutions. A more detailed discussion on Diffuser and Q-Learning for short-horizon problems is provided in Appendix \ref{app:theory_discuss}.

This naturally leads to the question: "Is it possible to devise an algorithm that excels in both settings?" Our analysis above suggests an affirmative response. We hypothesize that combining a sequence modeling method at the high level with a low-level value learning policy would be effective in both scenarios. To realize this concept, we introduce PlanDQ, a hierarchical planning framework for offline RL. Specifically, PlanDQ employs a diffusion-based planner, termed D-Conductor, for high-level planning while leveraging a Q-learning method, referred to as Q-Performer, for efficiently tackling low-level sub-tasks.

Our main contributions can be outlined as follows: 1) We propose a novel hierarchical planning method, PlanDQ, for offline decision-making; 2) Through empirical evaluation, PlanDQ is shown to either outperform or match the performance of existing state-of-the-art methods across a variety of tasks, demonstrating its effectiveness in both long-horizon and short-horizon settings. 3)  We provide evaluation reports for other possible orchestrating architectures, affirming the necessity and efficiency of PlanDQ. 4) Our empirical investigations further highlight that the guidance function within the Diffuser model is prone to converge towards sub-optimal solutions when applied to noisy data in short-horizon tasks, leading to inferior performance compared to value learning methods. 

\begin{table}[h]
  \centering
  \caption{Comparison between offline \textit{value function learning} and planning methods under different scenarios according to our experiments on the continuous control tasks.}
  \begin{adjustbox}{max width=\linewidth}
  
  \begin{tabular}{l|c|c}
  \toprule
Properties & Value function learning & Sequence modeling \\ \midrule
  Long-horizon Reasoning & \redxmark(Flat policy) & \greencheck \\
  Bellman Optimality & \greencheck (Short-horizon only)& \redxmark \\
  \bottomrule
  \end{tabular}
  \end{adjustbox}
  \label{table:intro_analysis}
\end{table}

\begin{figure*}[t]
\begin{center}
\centerline{\includegraphics[width=\linewidth]{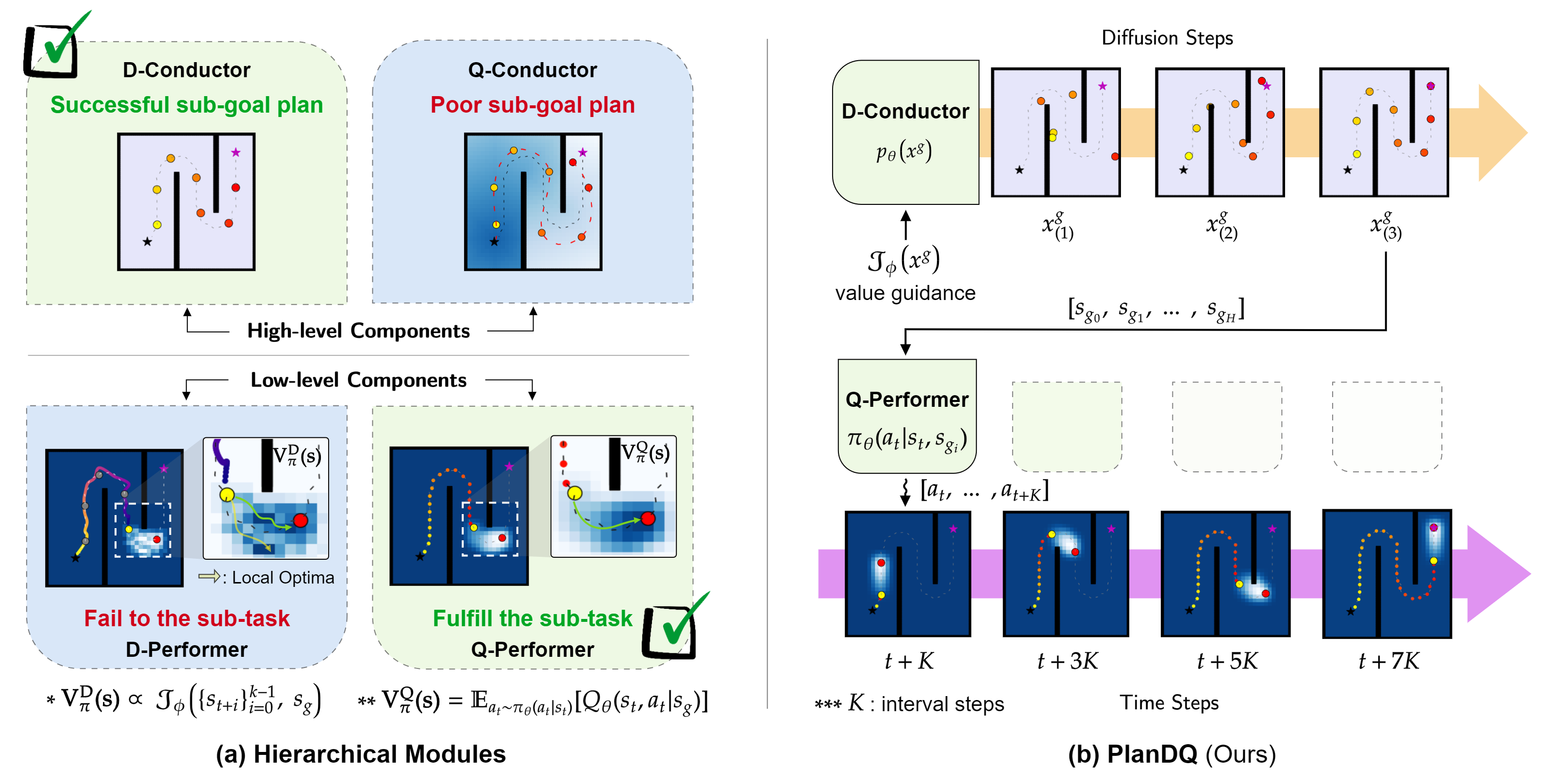}}
\caption{\textbf{Overview.}
To explore the optimal hierarchical planning architecture, we examine four hierarchical modules,\textit{ check} 1-(a), consisting of two conductors for high-level components and two performers for low-level components. And, we introduce a novel hierarchical planning architecture, called PlanDQ, which takes the D-conductor as a high-level component and the Q-performer as a low-level component, \textit{check }1-(b) The planning process of PlanDQ.}
\label{overview}
\vspace{-2em}
\end{center}
\end{figure*}

\section{Preliminaries}
\subsection{Offline Reinforcement Learning}
This work focuses on the \emph{offline reinforcement learning} (RL) framework. The offline RL, also known as batch RL, is a variant of RL where the learning process is conducted on a fixed static dataset $\mathcal{D} = \{(\rvs_t, \rva_t, \rvs_{t+1}, r_t)\}$ collected from behavior policies that we may not have access to. The general RL problem can be formulated as a Markov Decision Process (MDP): $\mathcal{M} = \{\mathcal{S}, \mathcal{A}, \mathcal{P}, \mathcal{R}, \gamma, \rvs_0\}$, with state space $\mathcal{S}$, action space $\mathcal{A}$, environment dynamics $\mathcal{P}(s' | s, a)$ representing probability of transitioning to state $\rvs'$ from state $\rvs$ after taking action $\rva$, reward function $\mathcal{R}(\rvs, \rva, \rvs')$ defining the expected immediate reward, $r$, after transitioning from state $\rvs$ to state $\rvs'$, discount factor $\gamma \in [0, 1]$, and initial state distribution $\rvs_0$. The goal of offline RL is to learn a policy $\pi_\theta(a |s)$ that maximizes the cumulative discounted reward purely from an offline dataset without any online interactions:
\begin{equation}
    \pi = \max_\pi \E_{a_t \sim \pi(\cdot | s_t)}\left[ \sum_{t=0}^\infty \gamma^t r(s_t, a_t) \right]\ .
\end{equation}
During this learning process, policy evaluation and policy improvement alternate: a parameterized critic network $Q_\phi(\rvs, \rva)$ is optimized to minimize the following temporal difference (TD) loss: $\mathcal{L}_\text{TD}(\phi) = \E_{(\rvs, \rva, \rvs') \sim \mathcal{D}}\left[ \bigl( r + \gamma \max_{\rva' \in \mathcal{A}} Q_{\hat{\phi}}(\rvs', \rva') - Q_\phi(\rvs, \rva) \bigr)^2 \right]$, where $Q_{\hat{\phi}}(\rvs, \rva)$, introduced to stabilize the training, is the target network that keeps a lagged copy of the weight of $Q_\phi(\rvs, \rva)$. The parameterized policy $\pi_\theta$ is trained to maximize the expected Q value: $\mathcal{L}_\pi = \E_{\rvs \in \mathcal{D}, \rva \sim \pi_\theta(\cdot \mid \rvs)}[Q_\phi(\rvs, \rva)]$.

While promising, offline RL faces significant challenges due to the function approximation errors encountered with out-of-distribution actions. To mitigate these issues, policy regularizization is often required to ensure the learned policy does not deviate too far from the behavior policy \citep{wang2022diffusion}.

\subsection{Diffusion Probabilistic Models}
Denoising Diffusion Probabilistic Model (DDPM)\cite{ho2020denoising} is a latent generative model that learns the data distribution $p_\theta(\rvx_0) \coloneqq \int p_\theta(\rvx_{0:M}) \mathrm{d}\rvx_{1:M}$, where $\rvx_1, \dots, \rvx_M$ are latent variables that share the same dimensionality as the data $\rvx_0$. After training, DDPM generates samples through an M-step iterative denoising process, starting from a Gaussian noise $\rvx_M \sim \mathcal{N}(\mathbf{0}, \mathbf{I})$:
\begin{align}
    &p_\theta (\rvx_0) = \int p(\rvx_M) \prod_{m=0}^{M-1} p_\theta (\rvx_m \mid \rvx_{m+1})\,\mathrm{d}\rvx_{1:M}\, \\
    &p_\theta (\rvx_m \mid \rvx_{m+1}) = \mathcal{N}(\rvx_m; \bm{\mu}_\theta (\rvx_{m+1}), \sigma_m^2 \mathbf{I})\ .
\end{align}

To learn a DDPM, the approximate posterior of the latents is given by a fixed diffusion process that gradually adds Gaussian noise to the data:
\begin{align}
    &q(\rvx_{1:M} | \rvx_0) \coloneqq \prod_{m=1}^M q(\rvx_m | \rvx_{m-1})\ ,\\
    &q(\rvx_m | \rvx_{m-1}) \coloneqq \mathcal{N}(\rvx_m; \sqrt{1 - \beta_m}\rvx_{m-1}, \beta_m \mathbf{I})\ ,
\end{align}
where $\beta_m$ is a pre-defined variance schedule $\sqrt{1-\beta_m} \rightarrow 0$ as $m \rightarrow \infty$, ensuring $\rvx_M$ would be a pure standard Gaussian noise. The training objective is to maximize the evidence lower bound defined as $\E_q\left[\ln \frac{p_\theta(\rvx_{0:M})}{q(\rvx_{1:M} | \rvx_0)}\right]$. 

In practice, the learnable mean $\mu_\theta(\rvx_m)$ is often parameterized as a linear combination of the latent $\rvx_m$ and the output of a noise-prediction U-net $\bm\epsilon$ \citep{unet}. The training objective is simplified \citep{ho2020denoising} as:
\begin{equation}
    \label{pre:diffusion_loss}
    \mathcal{L}_\theta = \E\left[ \lVert \bm\epsilon - \bm\epsilon_\theta(\rvx_m) \rVert^2 \right]\ .
\end{equation}

\subsection{Diffusion Models for Reinforcement Learning}
Current research applying diffusion models into Reinforcement Learning (RL) primarily focuses on three distinct applications: employing diffusion models as planners, as policy functions, and as tools for data synthesis. Given the scope of our work, we will briefly discuss the first two categories: using diffusion models as planners and as policy functions.

\textbf{Diffusion Planners.} Diffuser \cite{janner2022diffuser} was first proposed to learn a diffusion-based planner from the offline dataset. The key idea is to re-structure the trajectory of states and actions into a two-dimensional array:
\begin{equation}
    \label{pre:diffuser}
    \rvx =
    \begin{bmatrix}
        \rvs_0 & \rvs_1 & \dots & \rvs_T \\
        \rva_0 & \rva_1 & \dots & \rva_T
    \end{bmatrix}\ .
\end{equation}
A diffusion probabilistic model $p_\theta(\rvx)$ is then trained to model the data distribution. After training, Diffuser can plan a trajectory via sampling $\rvx \sim p_\theta(\rvx)$. To sample a plan for a specific task, Diffuser separately trains a guidance function $\mathcal{J}_\phi(\rvx)$ to predict the return $R(\rvx)$ of the trajectory $\rvx$ given a corrupted trajectory $\rvx_m$ as input. 
\begin{equation}
    \Ls(\phi) = \E_{\rvx, m, \bm{\epsilon}} \left[\lVert R(\rvx) - \mathcal{J}_\phi(\rvx_m) \rVert^2\right]\ ,
\end{equation}
Thus, a plan that tries to maximize the expected return can be sampled from the perturbed distribution, $\tilde{p}_\theta(\rvx) \propto p_\theta(\rvx)\exp(\mathcal{J}_\phi(\rvx))$, via classifier-guided sampling:
\begin{equation}
    \label{pre:guidance_sampling}
    \tilde{\bm{\mu}} \leftarrow \bm{\mu}_{\theta} (\rvx_{m+1}) + \omega \sigma_m^2 \nabla_{\rvx_m}\mathcal{J}_{\phi}(\rvx_m)|_{\rvx_m = \bm{\mu}_{\theta} (\rvx_{m+1})},
\end{equation}
where the hyperparameter $\omega$ controls the gradient scaling.

Follow-up works \citep{hdmi, hong2023offline, chen2024simple} have extended Diffuser with hierarchical structures and showed noticeable improvement on long-horizon tasks. The basic idea is to deploy Diffuser as a high-level planner to model a sub-goal sequence and train a low-level policy to follow the sub-goals. The trajectory consisted of sub-goals at the high level is formatted as:
\begin{equation}
    \label{pre:subgoal_seq}
    \rvx^g =
    \begin{bmatrix}
        \rvs_{g_0} & \rvs_{g_1} & \dots & \rvs_{g_H} \\
    \end{bmatrix}\ .
\end{equation}
, where $H$ denotes the subgoal planning horizon.

\textbf{Diffusion Policy Functions}
Diffusion Q-learning (DQL) \citep{wang2022diffusion} firstly explores the advantages of modeling the complex behavior policy with a diffusion model: $\pi_\theta(\rva \mid \rvs) = p_\theta(\rva_{0:N} \mid \rvs)$. To improve the policy, in addition to the diffusion reconstruction loss, the policy is also trained to maximize the expected return: $\mathcal{L}_\pi(\theta) = \E_{\rvs \sim \mathcal{D}, \rva_0 \sim \pi_\theta}[Q_\phi(\rvs, \rva_0)]$. The end sample, $\rva_0$, of the desnoising process $p_\theta(\rva_{0:N} \mid \rvs) = \mathcal{N}(\rva_N; \mathbf{0}, \mathbf{I})\prod_{i=1}^Np_\theta(\rva_{i-1} \mid \rva_{i}, \rvs)$ is used for RL evaluation. 

\section{Method}
In this section, we detail the design considerations behind our proposed methodology. Specifically, in Section \ref{method:hl-planner}, we describe the development of our high-level planner. Section \ref{method:ll-gc-policy} describes how to implement our goal-conditioned low-level policy. In Section \ref{method:model_variants}, we explore alternative hierarchical configurations and potential challenges. The overview structure of PlanDQ can be found in Figure \ref{overview}. 

\subsection{D-Conductor}\label{method:hl-planner}
Aligned with our hypothesis, we favor a sequence modeling approach for the high-level planner. Concurrently, our aim for the high-level module is to generate plausible sub-goals or sub-task abstraction (i.e., skills) that guide the low-level policy toward maximizing expected task returns. Various prior studies have explored different methods for implementing this high-level conductor \citep{sudhakaran2023skill, ma2023rethinking, hdmi, hong2023diffused}, within which the Hierarchical Diffuser (HD) proposed by \citealp{chen2024simple} standing out for its effectiveness and simplicity. To this end, following HD, we deploy diffusion models as our sub-goal planner, termed D-Conductor. Specifically, the sub-goal sequences are formatted with sampled intermediate states with fixed time interval $K$:
\begin{equation}
    \setlength\arraycolsep{2pt}
    \rvx^g =
    \begin{bmatrix}
        \rvs_{g_0} & \rvs_{g_1} & \dots & \rvs_{g_H} \\
    \end{bmatrix}\ =
    \begin{bmatrix}
        \rvs_{0} & \rvs_{K} & \dots & \rvs_{HK} \\
    \end{bmatrix}\ .
\end{equation}
Same as HD or Diffuser, a diffusion model $p_\theta(\rvx^g)$ is trained to minimize the simplified objective as in equation \ref{pre:diffusion_loss}.

While this generally works well for long-horizon tasks with sparse reward, to guide the plan towards maximizing the expected task return in a dense-reward setting, a separate guidance function $\mathcal{J}_\phi(\rvx^g)$ is required. The gradients of $\mathcal{J}_\phi$ guide the trajectory sampling procedure by modifying the means $\bm{\mu}$ of the denoising process as in equation \ref{pre:guidance_sampling}. Moreover, in this case, including the actions of the intermediate steps within trajectory representation helps the value function learning \citep{chen2024simple}. Here, we adopt the same strategy, and as a result, $\rvx^g$ becomes:
\begin{equation}
    \rvx^g =
    \begin{bmatrix}
        \rvs_0 & \rvs_K & \dots & \rvs_{HK} \\
        \rva_0 & \rva_K & \dots & \rva_{HK} \\
        \rva_1 & \rva_{K+1} & \dots & \rva_{HK+1} \\
        \vdots & \vdots & \ddots & \vdots \\
        \rva_{K-1} & \rva_{2K-1} & \dots & \rva_{(H+1)K-1} \\
    \end{bmatrix}\ .
\end{equation}

\begin{algorithm}[h]
\caption{PlanDQ: D-Conductor Training}
\label{alg:d_conductor}
\begin{algorithmic}[1]
    \STATE {\bfseries Input:} Initialized diffusion network $p_\theta$, value predictor $\mathcal{J}_\phi$, offline dataset $\mathcal{D}$, discount factor $\gamma$

    \FOR{each iteration}
        \STATE Sample a trajectory $\tau = \{(\rvs_t, r_t)\}^T$
        \STATE Extract state sequence $\{\rvs_t\}^{HK}$ to form $\rvx^q$
        \STATE Compute return $R(\tau)=\sum_{t}^T \gamma^t r_t$
        \STATE Sample $m \sim Uniform(\{1, \dots, M\})$, noise $\bm{\epsilon} \sim \mathcal{N}(\bm{0}, \bm{1})$ 
        \STATE Update $p_\theta$ by minimizing Equation \ref{pre:diffusion_loss}
        \STATE Update $\mathcal{J}_\phi$ by minimizing Equation \ref{pre:guidance_sampling}
    \ENDFOR
\end{algorithmic}
\end{algorithm}

\subsection{Q-Performer}\label{method:ll-gc-policy}
Continuing with our hypothesis, we opt for a value function learning method at the low level. This preference is due to the nature of sub-tasks encountered by the low-level policy are typically short horizons and dense rewards (augmented with intrinsic rewards). Value function learning methods can theoretically achieve Bellman optimality through efficient credit assignment. However, directly applying off-policy RL methods to offline learning will suffer from the extrapolation error problem \citep{fujimoto2019off}. Drawing inspiration from recent breakthroughs that utilize expressive diffusion models to capture complex behavior policies, we propose employing a goal-conditioned diffusion model as our low-level policy. Specifically, given the current state $\rvs$ and sub-goal $\rvs_g$ from the D-Conductor, we define our Q-performer:
\begin{equation}
    \pi_\theta(\rva | \rvs, \rvs_g) = p_\theta(\rva^M)\prod_{m=1}^M p_\theta(\rva_{m-1} | \rva_{m}, \rvs, \rvs_g)\ .
\end{equation}
The objective for policy regularization can be written by:
\begin{equation}
    \mathcal{L}_d(\theta) = \E_{m, \epsilon, \rva^0, \rvs, \rvs_g} \left[ \lVert \bm\epsilon - \bm\epsilon_{\theta^Q} (\rva^m, m; \rvs, \rvs_g)\rVert^2\right]\ ,
\end{equation}
where $\bm\epsilon_{\theta^Q}$ is implemented as a 3-layer MLP with Mish activations; details of implementation can be found in the Appendix \ref{app:implementation}.

To evaluate the policy, the goal-conditioned action-value function is learned with a one-step RL as also proposed by \citep{gulcehre2021regularized,brandfonbrener2021offline, pmlr-v37-schaul15} for offline RL as a stable way of learning Q-functions, minimizing the TD error $\mathcal{L}_\text{TD}(\phi)$:
\begin{equation}
    \label{method:q_value_loss}
    \E_{(\rvs, \rva, \rvs') \sim D}\left[ \bigl(r^{\rvs_g} \! + \! \gamma Q_{\phi^\prime}(\rvs' ,\rva', \rvs_g) \! - \! Q_\phi(\rvs, \rva, \rvs_g) \bigr)^2\right]\!, 
\end{equation}
    
where $\rva' \sim \pi_\theta(\rva \mid \rvs)$, and the goal-conditioned reward $r(\rvs, \rva, \rvs_g)$ is defined as the combination of external task reward and intrinsic goal-reaching reward: $r^{\rvs_g} = r^{\text{ext}}(\rvs, \rva) + r^{\text{intr}}(\rvs, \rva, \rvs_g)$. In Section \ref{exp:intr_reward}, we conduct an ablation study to examine the impact of intrinsic rewards on our model's performance. Further details regarding the intrinsic reward are available in the Appendix \ref{app:implementation}.

To improve the policy, we minimize the following loss function with hyperparameter $\alpha$ controlling the trade-off between policy extraction and policy improvement as done by \cite{kumar2020conservative}:
\begin{equation}
    \label{method:policy_improve}
    \mathcal{L}_q(\theta) = - \alpha \E_{\rvs  \sim D, \rva \sim \pi_\theta} \left[ Q_\phi(\rvs, \rva, \rvs_g) \right]\ .
\end{equation}
Thus, the final objective for the policy training is given by $\mathcal{L}_\pi(\theta) = \mathcal{L}_d(\theta) + \mathcal{L}_q(\theta)$.
In practice, the training of the Q-Performer can proceed concurrently with that of the D-Conductor, as outlined in Algorithm \ref{alg:d_conductor} and Algorithm \ref{alg:q_performer}. Empirically, we found that randomly sampling the sub-goal for each state $\rvs_t$ from the range $[t, t+\Delta*K]$, where $\Delta \sim Geom(p)$, with $p=0.2$, makes training stable.

\subsection{Orchestrating Family}\label{method:model_variants}
The advantage of hierarchical structures for long-horizon tasks in offline RL is well known \citep{Ajay2020iris, pertsch2020long, rosete2023latent}. Therefore, it is worthwhile to investigate alternative hierarchical configurations. Namely PlanQQ, PlanQD, and PlanDD, to further investigate the potential of hierarchical configurations.

In PlanQQ, diverging from PlanDQ, a Q-learning-based conductor, referred to as a Q-Conductor, is utilized at the high level to function as a sub-goal generator: $\pi_\theta(\rvs_{t+K} | \rvs_{t})$. This Q-Conductor produces a state that is $K$ steps ahead of the current state $\rvs_t$ as the sub-goal \citep{park2023hiql}. The Q value function is optimized using the loss function $\mathcal{L}_\text{TD}(\phi) = \E_{(\rvs_t, \rvs_{t+K}) \sim D}\left[ \bigl( R^K + \gamma Q_{\phi^\prime}(\rvs_{t+K} ,\rvs') - Q_\phi(\rvs_t, \rvs_{t+K}) \bigr)^2\right]$, where $R^K = \sum_{i=0}^{K-1}r_{t+i}$. Similar to PlanDQ's Q-Performer, the high-level policy aims to minimize diffusion and policy improvement loss.

PlanQD introduces a variation from PlanQQ by incorporating a Diffuser at the low level to model the sub-trajectory linking state $s_t$ to state $s_{t+K}$, including actions, as in equation \ref{pre:diffuser}. The first action $\rva_0$ is executed during RL evaluation.

In PlanDQ, high-level and low-level planners are instantiated as Diffusers, which has been proposed by \citealp{chen2024simple}. A detailed comparison of the experimental performance of these model variants, including PlanDD, is provided in Section \ref{exp:ablation-study}.

\begin{algorithm}[h]
\caption{PlanDQ: Q-Performer Training}
\label{alg:q_performer}
\begin{algorithmic}[1]
    \STATE {\bfseries Input:} Initialized policy network $\pi_\theta$, Q-value network $Q_\phi$, and target Q-value network $Q_{\phi'}$, offline dataset $\mathcal{D}$
    
    \FOR{each iteration}
        \STATE Sample a mini-batch $\mathcal{B} = \{(\rvs_t, \rva_t, r_t, \rvs_{t+1})\}$
        \STATE Sample a sub-goal $\rvs_g$ for each $\rvs_t$ 
        \STATE Update $Q_\phi$ by minimizing Equation \ref{method:q_value_loss}
        \STATE Update policy by minimizing $\mathcal{L}_\pi(\theta)$
    \ENDFOR
\end{algorithmic}
\end{algorithm}

\section{Experiment}

Our experiment section first presents our main results on the standard D4RL \citep{d4rl} benchmarks. We also include tasks with extended horizons, specifically designed to assess the long-horizon reasoning capabilities of the model. Following this, we conduct an analysis using a simplified OpenMaze2D environment, offering insights into the reasons behind the superior performance of the Q-learning-based methods compared to value-guided sequence modeling approaches. We end our experiment section with a thorough analysis of our proposed method. 
\subsection{Experimental Setup}
\textbf{Benchmarks.}  We first briefly introduce the tasks we used to evaluate our method. The AntMaze suite, known for its challenging long-horizon navigation tasks, is a task where the goal is to control an 8-DoF Ant to reach a pre-set goal from its initial position. Beyond the standard levels of AntMaze included in the D4RL benchmark, our evaluation extends to AntMaze-Ultra \citep{jiang2023efficient}, which introduces a larger maze environment. The Kitchen from D4RL and Calvin \citep{mees2022calvin} are two long-horizon manipulation tasks. In these two tasks, the agent is required to complete a set of subtasks. The agent will receive a reward of $1$ upon the completion of each subtask. The Gym-MuJoCo suite consists of control tasks designed for short-horizon and dense-reward. 

\textbf{Baselines.} We consider different classes of baselines that perform well in each domain of tasks. CQL (Conservative Q-Learning, \citealp{kumar2019stabilizing}) is a Q-value constraint offline-RL method. IQL(Implicit Q-Learning, \citealp{kostrikov2022offline}) is a policy regularization-based method. TT (Trajectory Transformer, \citealp{janner2021offline}) utilizes sequence modeling, which leverages the Transformer \citep{transformer} to model the entire trajectories. In contrast to TT, Diffuser \citep{janner2022diffuser} and DD \citep{decision_diffuser} are two planning-based methods, employing a diffusion probabilistic model \citep{ho2020denoising} to model the trajectory distribution. DQL (Diffusion-QL, \citealp{wang2022diffusion}), meanwhile, utilizes the same diffusion probabilistic model but focuses on learning the behavior policy. Within the realm of hierarchical learning, we consider HIQL (Hierarchical implicit Q-learning, \citealp{park2023hiql}), which extends IQL to hierarchical framework. HDMI \citep{hdmi}, DTAMP \citep{hong2023diffused}, and HD \citep{chen2024simple} are hierarchical diffusion-based planners. For the purpose of streamlined comparison, we categorize these methods into two groups: flat learning methods and hierarchical learning methods. The source of baseline performances are detailed in Appendix \ref{app_sec:source_baseline}.
\begin{table*}[t]
\centering
\caption{\textbf{Long-horizon Navigation and Manipulation.} PlanDQ combines the benefits of both diffusion-based planning and value-based policy learning, achieving the best performance across all tasks. PlanDQ results are averaged over $100$ random planning seeds for AntMaze tasks, and $15$ random planning seeds for Kitchen and Calvin tasks.}
\begin{adjustbox}{max width=0.95\linewidth}
\label{table:long_horizon}
\begin{tabular}{llrrrrrrrrr}

\toprule 
\multicolumn{2}{c}{\multirow{2}{*}{\textbf{Environment}}} & \multicolumn{4}{c}{\textbf{Flat Learning Methods}} & \multicolumn{5}{c}{\textbf{Hierarchical Learning Methods}} \\ \cmidrule(r){3-6} \cmidrule(l){7-11}
& &\textbf{IQL}      &\textbf{DQL}     &\textbf{Diffuser}  &\textbf{DD}  & \textbf{HIQL}    & \textbf{HDMI}     & \textbf{DTAMP}  & \textbf{HD} & \textbf{PlanDQ}(ours)\\ \midrule
AntMaze & Medium &$70.0$ & $90.0 \! \pm \! 3.0$ & $31.9 \! \pm \! 5.1$ &$0.0$ &$86.8 \! \pm \! 4.6$ &$-$ & $88.7 \! \pm \! 2.5$  &$88.7 \! \pm \! 8.1$ & $\mathbf{93.0} \! \pm \! 2.6$\\
AntMaze & Large  &$47.5$ & $64.0 \! \pm \! 4.8$ & $0.0 \! \pm \! 0.0$ &$0.0$ &$\mathbf{88.2} \! \pm \! 5.3$ &$-$ & $78.0 \! \pm \! 3.1$  &$83.6 \! \pm \! 5.8$  & $86.0 \! \pm \! 3.5$\\ 
AntMaze & Ultra  &$21.6$ & $11.0 \! \pm \! 3.0$ & $0.0 \! \pm \! 0.0$ &$0.0$ &$52.9 \! \pm \! 17.4$ &$-$ &$-$ &$53.3 \! \pm \! 12.9$  & $\mathbf{70.0} \! \pm \! 4.5$\\ \midrule
\multicolumn{2}{c}{\textbf{AntMaze Average}}   &$46.4$ &$55.0$ &$10.6$ &$0.0$ &$76.0$  &$-$   &$-$    & $75.2$   &$\mathbf{83.0}$\\ \midrule \midrule
Kitchen & Mixed     &$51.0$ & $62.6$ & $50.0 \! \pm \! 8.8$ &$65.0$ &$67.7 \! \pm \! 6.8$ & $69.2 \! \pm \! 1.8$    &$\mathbf{74.4} \! \pm \! 1.4$ & $71.7 \! \pm \! 2.7$    & $71.7 \! \pm \! 2.7$ \\
Kitchen & Partial   &$46.3$ & $60.5$ & $56.2 \! \pm \! 5.4$ &$57.0$  &$65.0 \! \pm \! 9.2$ & $-$                     &$63.4 \! \pm \! 8.8$ & $73.3 \! \pm \! 1.4$    & $\mathbf{75.0} \! \pm \! 7.1$ \\ \midrule \midrule
\multicolumn{2}{c}{\textbf{Kitchen Average}}   &$48.7$ &$61.6$ &$53.1$ &$61.0$  &$66.4$   &$-$        & $68.9$          &$72.5$                       & $\mathbf{73.4}$  \\ \midrule \midrule
Calvin & & $7.8$ & $N/A$ & $37.5 \! \pm \! 9.5$ &$40.0 \! \pm \! 16.6$ & $43.8 \! \pm \! 39.5$ &$-$ & $-$  &$31.7 \! \pm \! 8.2$ & $\mathbf{45.0} \! \pm \! 19.8$\\ \midrule
\bottomrule

\end{tabular}
\end{adjustbox}
\end{table*}

\begin{table*}[t]
\small
\centering
\caption{\textbf{Offline Reinforcement Learning.} PlanDQ achieves the best overall performance among hierarchical learning methods. PlanDQ results are averaged over $5$ random planning seeds. Following~\citet{kostrikov2022offline}, we emphasize in bold scores within $5\%$ of maximum.}
\begin{adjustbox}{width=\linewidth}
\label{table:short_horizon}
\begin{tabular}{llrrrrrrrrr}
\toprule 
\multicolumn{2}{c}{\multirow{2}{*}{\textbf{Environment}}} & \multicolumn{5}{c}{\textbf{Flat Learning Methods}} & \multicolumn{4}{c}{\textbf{Hierarchical Learning Methods}} \\ \cmidrule(r){3-7} \cmidrule(l){8-11}
\multicolumn{2}{c}{ \textbf{Gym Tasks} } & \textbf{CQL} & \textbf{IQL} & \textbf{DD} & \textbf{TT} & \textbf{DQL}  &\textbf{DTAMP} &\textbf{HDMI} &\textbf{HD} & \textbf{PlanDQ}(ours) \\ \midrule
Med-Expert & HalfCheetah & $91.6$ & $86.7$ & $90.6 \! \pm \! 1.3$ & $\mathbf{95.0}$ & $\mathbf{96.8} \! \pm \! 0.3$ &$88.2$ & $92.1 \! \pm  \!1.4$ & $92.5 \! \pm \! 0.3$ & $\mathbf{95.4} \! \pm 0.3$\\
Med-Expert & Hopper  & $105.4$ & $91.5$ & $\mathbf{111.8} \! \pm \! 1.8$ & $\mathbf{110.0}$ & $\mathbf{111.1} \! \pm \! 1.3$ & $109.4$ &$\mathbf{113.5} \! \pm \! 0.9$ & $\mathbf{115.3} \! \pm \! 1.1$ & $\mathbf{111.6} \! \pm \! 0.2$\\
Med-Expert & Walker2d  & $\mathbf{108.8}$ & $\mathbf{109.6}$ & $\mathbf{108.8} \! \pm \! 1.7$ & $101.9$  & $\mathbf{110.1} \! \pm \! 0.3$ & $108.2$ & $\mathbf{107.9} \! \pm \! 1.2$ & $\mathbf{107.1} \pm 0.1$ &$\mathbf{110.2} \! \pm \! 0.1$ \\ \midrule 
\multicolumn{2}{c}{ \textbf{Med-Expert Average} }  & $101.9$ & $96.0$ & $103.7$ & $102.3$ & $\mathbf{106.0}$ & $101.9$ & $104.5$ & $105.0$ &$\mathbf{105.7}$\\ \midrule \midrule
Medium & HalfCheetah  & $44.0$ & $47.4$ & $49.1 \! \pm \! 1.0$ & $46.9$ & $\mathbf{51.1} \! \pm \! 0.5$ & $47.3$ &$48.0 \! \pm \! 0.9$ & $46.7 \! \pm \! 0.2$ &$\mathbf{50.2} \! \pm \! 0.1$\\
Medium & Hopper  & $58.5$ & $66.3$ & $79.3 \! \pm \! 3.6$ & $61.1$ & $90.5 \! \pm \! 4.6$ & $80.7$ & $76.4 \! \pm \! 2.6$& $\mathbf{99.3} \! \pm \! 0.3$ &$\mathbf{96.9} \! \pm \! 1.3$\\
Medium & Walker2d  & $72.5$ & $78.3$ & $82.5 \! \pm \! 1.4$ & $79.0$ & $\mathbf{87.0} \! \pm \! 0.9$ & $82.7$ &$79.9 \! \pm \! 1.8$ & $84.0 \! \pm \! 0.6$ &$\mathbf{86.5} \! \pm \! 0.2$ \\ \midrule 
\multicolumn{2}{c}{ \textbf{Med-Expert Average} }  & $58.3$ & $64.0$ & $70.3$ & $62.3$ & $76.2$ & $70.3$ & $68.1$ & $76.7$ &$\mathbf{77.9}$\\ \midrule \midrule
Med-Replay & HalfCheetah & $45.5$ & $44.2$ & $39.3 \! \pm \! 4.1$ & $41.9$ & $\mathbf{47.8} \! \pm \! 0.3$ & $42.6$ & $44.9 \! \pm \! 2.0$ & $38.1 \! \pm \! 0.7$ &$\mathbf{47.6} \! \pm \! 0.1$\\
Med-Replay & Hopper  & $95.0$ & $94.7$ & $\mathbf{100.0} \! \pm \! 0..7$ & $9 1 . 5$ & $\mathbf{101.3} \! \pm \! 0.6$ & $100.0$ & $\mathbf{99.6} \! \pm \! 1.5$& $ 94.7 \! \pm \! 0.7$ &$\mathbf{101.4} \! \pm \! 0.6$ \\
Med-Replay & Walker2d  & $77.2$ & $73.9$ & $75.0 \! \pm \! 4.3$ & $82.6$ & $\mathbf{95.5} \! \pm \! 1.5$  & $79.5$ & $80.7 \! \pm \! 2.1$ & $84.1 \! \pm \! 2.2$ &$\mathbf{94.0} \! \pm \!0.2$  \\ \midrule 
\multicolumn{2}{c}{ \textbf{Med-Replay Average} }  & $72.6$ & $70.9$ & $71.4$ & $72.0$ & $\mathbf{81.5}$ & $74.0$ & $75.1$ & $72.3$ &$\mathbf{81.0}$\\ \midrule
\bottomrule
\end{tabular}
\end{adjustbox}
\end{table*}

\subsection{Long-horizon Navigation and Manipulation}
We first evaluate our method on three tasks (AntMaze, Kitchen, Calvin) designed explicitly for long-horizon reinforcement learning. We assess performance based on the ability to achieve task-specific targets, such as reaching the target position in AntMaze or completing all four subtasks in Kitchen and Calvin. As indicated in Table \ref{table:long_horizon}, PlanDQ outperforms other methods regarding average task performance, showing a notable $8\%$ improvement over the most robust baseline, HIQL, in the AntMaze task. Additionally, hierarchical learning methods generally outperform flat learning methods in the long-horizon domains. This superiority is attributed to enhanced value estimation, as exemplified in HIQL, or to more efficient long-horizon sequence modeling, as evidenced in HDMI, DTAMP, HD, and our proposed method. Notably, PlanDQ's comparison with DTAMP and HD underscores the effectiveness of utilizing a Q-learning-based diffusion policy at the low level.

\subsection{Short-horizon Controlling}
We further demonstrate that utilizing a Q-performer at the low level shows enhanced performance over other hierarchical learning methods in short-horizon tasks with dense rewards. As depicted in Table \ref{table:short_horizon}, in contrast to the long-horizon tasks discussed previously, nearly all hierarchical learning methods, except PlanDQ, underperform relative to the strongest flat learning method. We exclude HIQL from this comparison as it is specifically designed for goal-conditioned tasks and unsuitable for scenarios without a defined goal. PlanDQ is the only hierarchical learning method that matches the performance of the leading flat learning baselines. Notably, against HD, PlanDQ exhibits a significant improvement of $12\%$ on the Medium-Replay dataset, which consists of a noisy buffer accumulated from a policy trained to the medium agent's level of performance. An overall comparison can be found in Figure \ref{benchoverview}, where the performance of the baselines is normalized relative to that of PlanDQ. The following section delves deeper into this improvement, and we hope this can offer experimental insights into the underlying reasons for this advancement.

\subsection{Analysis and Ablation Study}\label{analysis}\label{exp:ablation-study}
\begin{figure}[h]
\begin{center}
\centerline{\includegraphics[width=0.7\linewidth]{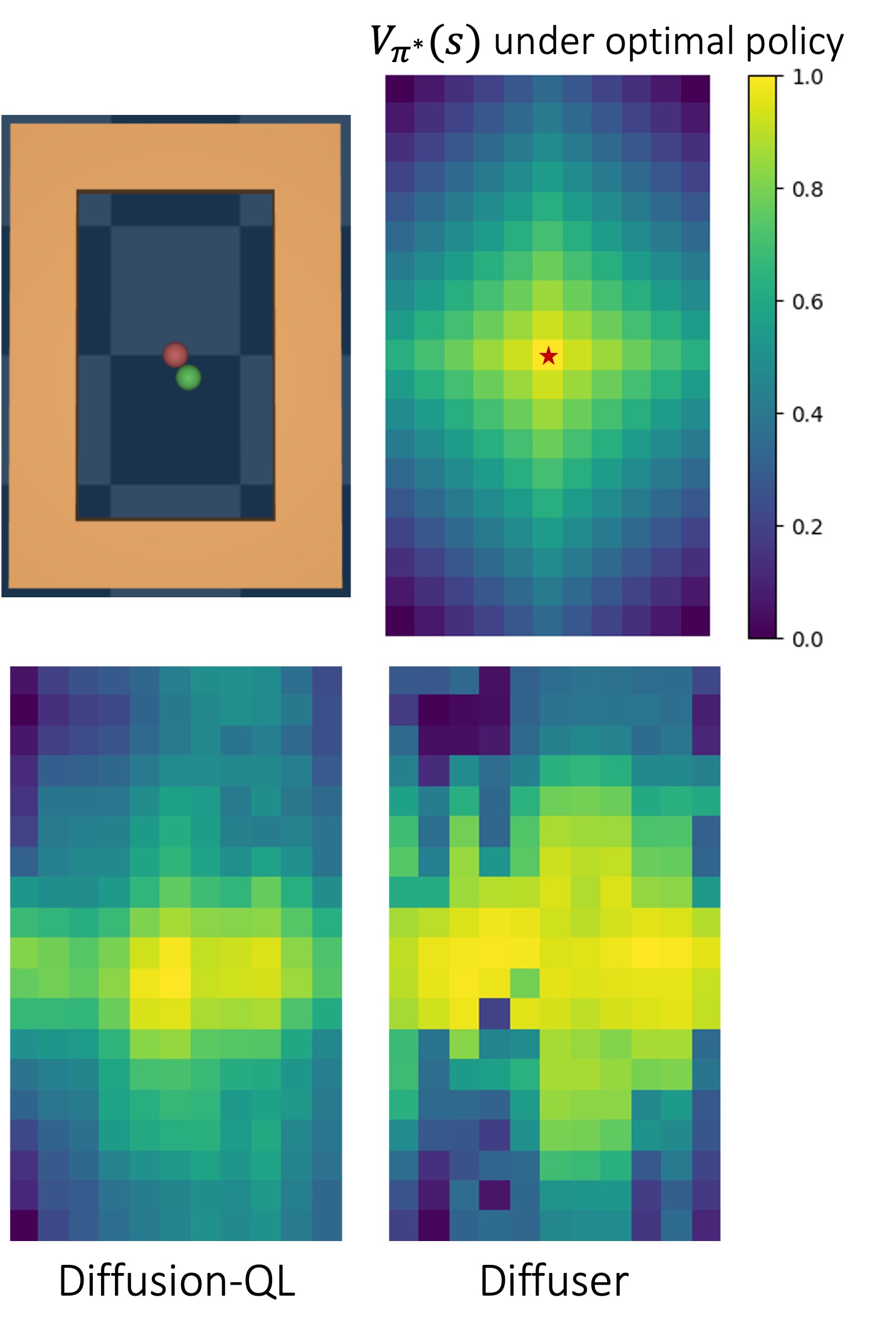}}
\vspace{-1em}
\caption{\textbf{Value Estimation Comparison.} Compared with Diffusion-QL, Diffuser learns a more noisy value function.}
\vspace{-2em}
\label{fig:exp_toy}
\end{center}
\end{figure}

\textbf{Diffuser model's value prediction converges to a local optimum in a toy task.}\label{exp:toy}
To elucidate the reasons for PlanDQ's improvement over HD on a noisy dataset, we conducted a simplified experiment in the OpenMaze environment, designed to replicate the short-horizon and dense reward setting at a lower level. This environment, showcased in Figure \ref{fig:exp_toy}, involves guiding a point-mass agent to a fixed goal located at the center of a small open maze. In contrast to a sparse reward setting, the agent receives a reward at each timestep, calculated as the negative exponentiated distance from the goal. To simulate noisiness, we generated a sub-optimal dataset using a controller, randomly selecting goal locations while deliberately avoiding the true central goal.

In this setup, we compared Diffusion-QL (DQL) and the Diffuser, the underlying low-level policies in PlanDQ and HD, respectively. The results, presented in Table \ref{table:exp_toy}, demonstrate that DQL outperforms Diffuser by a margin of $20\%$. To delve deeper into the causes of this performance discrepancy, we plotted the estimated value maps for both methods alongside the value map from an optimal policy, as seen in Figure \ref{fig:exp_toy}. The visual comparison reveals that the learned value function in DQL correctly predicts higher values for states near the center. In contrast, the value predictor in Diffuser erroneously assigns higher values to states away from the center. As a consequence, when guided by its value function, Diffuser tends to lead the agent in suboptimal directions, resulting in less effective behaviors in the task. A potential explanation for this local optimality in Diffuser's value prediction is its learning approach, which involves regressing to the cumulative rewards of trajectories without the Bellman updates. When the horizon is short, this approach tends to introduce errors in value prediction, adversely affecting performance.

\begin{table}[h]
    \centering
    \caption{RL performance on the toy OpenMaze task.}
    \begin{adjustbox}{max width=0.65\linewidth}
       \begin{tabular}{crr}
        \toprule
        \textbf{Environment} &  \textbf{DQL} & \textbf{Diffuser} \\ \midrule
         OpenMaze   &$\mathbf{90.4} \! \pm \! 1.9$ & $74.9 \! \pm \! 1.2$\\  \bottomrule
    \end{tabular} 
    \end{adjustbox}
    \label{table:exp_toy}
\end{table}

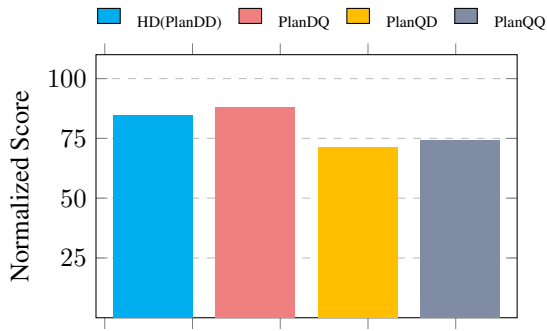
\begin{figure}[h]
    \centering
    \begin{tikzpicture}
    \pgfplotsset{
            scale only axis,
        }
        \begin{axis}[ybar=2pt,
        ymin=00,
        ymax=110,
        ytick={25, 50, 75, 100},
        symbolic x coords={HD(PlanDD), PlanDQ, PlanQD, PlanQQ},
        xticklabel=\empty,
        bar width=30pt,
        width=5.6cm,
        height=3.5cm,
        ymajorgrids=true,
        grid style=dashed,
        enlarge x limits=3.5,
        ylabel near ticks,
        xlabel near ticks,
        legend cell align=left,
        legend style={draw=none, fill=none, text opacity = 1,column sep=0.15cm, {at={(1.1,1.2)}}, font=\tiny, legend columns=-1},
        legend image code/.code={
        \draw [#1] (0cm,0cm) rectangle (0.3cm,0.2cm); },
        ylabel={Normalized Score},
        xticklabel style={yshift=3pt},
        ]
        \legend{HD(PlanDD), PlanDQ, PlanQD, PlanQQ}
        \addplot[ybar, fill=cyan, mark=none, draw=none,
        ]
        coordinates {
        (HD(PlanDD),84.6)
        };
        \addplot[ybar, fill=lightcoral, mark=none, draw=none,
        ]
        coordinates {
        (PlanDQ,88.2)
        };
        \addplot[ybar, fill=fluorescentorange, mark=none, draw=none,
        ]
        coordinates {
        (PlanQD,71.1)
        };
        \addplot[ybar, fill=bluegray, mark=none, draw=none,
        ]
        coordinates {
        (PlanQQ,74.3)
        };
        \end{axis}
    \end{tikzpicture}
    \caption{\textbf{Model Exploration on the Gym-MuJoCo.} PlanDQ achieves the best averaged performance over all the model variants.}
    \label{fig:ablate_model_variants}
\end{figure}

\textbf{Q-conductor is not stable at the high-level.}
We verified the benefits of utilizing a Q-performer at the low-level, and it is intriguing to consider whether a Q-learning-based method could similarly enhance the high-level planner. To explore this, we conducted an ablation analysis on various configurations within our hierarchical framework. Specifically, we examined different combinations of diffusion planning and value-based learning for both the high-level conductor and low-level performer. These combinations include PlanDD, which utilizes diffusion planning at both levels; PlanQD, with Q-learning at the high level and diffusion planning at the low level; and PlanQQ, featuring Q-learning at both levels. The normalized scores for these models on the Gym-MuJoCo tasks are presented in Figure \ref{fig:ablate_model_variants}.
The results indicate that employing Q-learning at the high level leads to a decrease in performance on the dense reward tasks. A plausible explanation for this is that the value functions are approximations of expected return with respect to the learning policy, the reward term $R^K$ is collected from the behavior policy, without correction, this makes the Q value function hard to learn with the naive Bellman update. To the best of our knowledge, there is no simple fix within the scope of our problem setting (i.e., offline RL). We will leave this interesting direction for future study.


\textbf{Q-performer outperforms other behavior cloning variants.} The low-level performer solves sub-tasks output from the high-level conductor. A goal-conditioned behavior cloning (BC) approach also appears capable of accomplishing this task. To evaluate the necessity and effectiveness of a diffusion-based Q-learning approach, we conduct experiments on the short-horizon Gym-MuJoCo and long-horizon sparse reward AntMaze and Kitchen tasks, comparing against variants with a pure BC performer as well as an agent trained with TD3+BC at the low-level  \citep{fujimoto2021minimalist}. The results are detailed in Table \ref{table:ablation_bc_gym} and Table \ref{table:ablation_bc_addtional} in the Appendix. The findings indicate that Q-learning performers outperform pure BC performer on these tasks. This superiority may stem from the additional guidance the Q-values provide about the external tasks. Moreover, in comparison with TD3+BC, PlanDQ achieved superior performance due to a more expressive diffusion-based policy.
\begin{table}[h!]
\small
\centering
\caption{\textbf{Ablation Study on Low-Level Performer.} PlanDQ outperforms variants with behavior cloning performers on the Gym-MuJoCo medium replay dataset. Results are averaged over 5 planning seeds.}
\begin{adjustbox}{width=0.9\linewidth}
\label{table:ablation_bc_gym}
\begin{tabular}{cccr}
\toprule 
\textbf{Tasks}         & \textbf{BC}          & \textbf{TD3+BC}          & \textbf{PlanDQ}         \\ \hline
HalfCheetah   & $37.6 \! \pm \! 0.4$ & $46.2 \! \pm \! 0.9$ & $\mathbf{46.7} \! \pm \! 0.1$ \\ 
Hopper   & $23.6 \! \pm \! 3.1$ & $96.8 \! \pm \! 1.4$  & $\mathbf{101.4} \! \pm \! 0.6$ \\  
Walker2d   & $18.8 \! \pm \! 2.0$ & $78.9 \! \pm \! 1.2$  & $\mathbf{94.0} \! \pm \! 0.2$ \\  \hline
\textbf{Average} & $26.7$         & $74.0$         & $\mathbf{81.0}$         \\ \hline
\end{tabular}
\end{adjustbox}
\end{table}

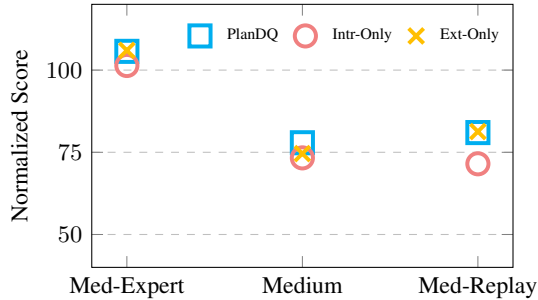
\begin{figure}[h]
    \centering
    \begin{tikzpicture}
    \pgfplotsset{
            scale only axis,
        }
        \begin{axis}[
        ymin=40,
        ymax=120,
        ytick={50, 75, 100},
        xtick={Med-Expert, Medium, Med-Replay},
        symbolic x coords={Med-Expert, Medium, Med-Replay},
        width=5.6cm,
        height=3.5cm,
        ymajorgrids=true,
        grid style=dashed,
        ylabel near ticks,
        xlabel near ticks,
        label style={font=\footnotesize},
        tick label style={font=\small},
        legend cell align=left,
        legend style={draw=none, fill=none, text opacity = 1,column sep=0.15cm, {at={(1.0,0.95)}}, font=\tiny, legend columns=-1},
        ylabel={Normalized Score},
        ]
        \legend{PlanDQ, Intr-Only, Ext-Only}
        \addplot[only marks, mark size=4, color=cyan, mark=square, draw=none, mark options={fill=cyan, line width=1.5pt}]
        coordinates {
        (Med-Expert,105.7)(Medium, 77.9)(Med-Replay, 81.0)
        };
        \addplot[only marks, mark size=4, color=lightcoral, mark=o, draw=none, mark options={fill=cyan, line width=1.5pt}
        ]
        coordinates {
        (Med-Expert,101.4)(Medium, 73.3)(Med-Replay, 71.5)
        };
        \addplot[only marks, mark size=4, color=fluorescentorange, mark=x, draw=none, mark options={fill=cyan, line width=1.5pt}
        ]
        coordinates {
        (Med-Expert,106.0)(Medium, 74.6)(Med-Replay, 81.3)
        };
        \end{axis}
    \end{tikzpicture}
    \caption{\textbf{PlanDQ with different Low-level rewards scheme on the Gym-MuJoCo.} The combination of external rewards with intrinsic rewards achieves the best-averaged performance over different qualities of dataset.}
    \label{fig:ablate_rewards}
\end{figure}

\textbf{External rewards together with intrinsic rewards achieve the best performance.}\label{exp:intr_reward} As outlined in earlier sections, our low-level Q-Performer is designed to maximize both cumulative external rewards and intrinsic rewards. In this analysis, we evaluate the effectiveness of this approach by comparing PlanDQ against two variants: Intr-Only, where the low-level Q-Performer is trained exclusively on intrinsic rewards, and Ext-Only, which is trained solely on external rewards. As illustrated in Figure \ref{fig:ablate_rewards}, PlanDQ outperforms the other configurations across all dataset qualities. The Intr-Only variant shows weaker performance on the Med-Replay dataset, potentially due to the sub-optimality of the sub-goals generated by the high-level Diffuser. Because it is challenging for Diffuser to learn an optimal value predictor with noisy dataset, as demonstrated in our toy experiments. While the performance of Ext-Only is comparable to that of PlanDQ, PlanDQ notably excels on the Medium dataset.

\textbf{PlanDQ performs reasonably well with different sub-goal time interval $K$.} Lastly, we assess the robustness of PlanDQ in relation to varying sub-goal time interval $K$. Intuitively, a larger $K$ may lead to information loss at the high level, resulting in suboptimal sub-goal planning, and could also decrease the efficiency of Q-learning at the low level. Conversely, a smaller $K$ might negate the advantages of a broader receptive field provided by the hierarchical structure \citep{chen2024simple}. We conducted experiments on the Gym-MuJoCo and AntMaze tasks with various $K$ values and, due to space constraints, report the Gym-MuJoCo results in Table \ref{table:ablation_k}. For the AntMaze results, please refer to the corresponding table in the Appendix. Specifically, we tested $K \in \{2, 4, 8\}$ on the Gym-MuJoCo tasks and $K \in \{15, 30, 45\}$ on the AntMaze tasks. PlanDQ generally performs well across different $K$ values, albeit with a slight performance drop at $K=8$.

\begin{table}[h!]
\small
\centering
\caption{\textbf{Ablation Study on Sub-goal Time Interval Step $K$.} PlanDQ performs reasonably well with different jumpy steps $K$. Results are averaged over 5 planning seeds.}
\begin{adjustbox}{width=\linewidth}
\label{table:ablation_k}
\setlength{\tabcolsep}{4.5pt}
\begin{tabular}{llccr}
\toprule 
\multicolumn{2}{c}{\textbf{Environment}}  & K$2$         & K$4$          & K$8$          \\ \hline
Med-Exp & HalfCheetah   & $\mathbf{96.1} \! \pm \! 0.1$ & $95.4 \! \pm \! 0.3$ & $94.8 \! \pm \! 0.6$ \\ 
Med-Exp & Hopper   & $\mathbf{112.9} \! \pm \! 0.2$ & $111.6 \! \pm \! 0.2$  & $110.5 \! \pm \! 0.8$ \\  
Med-Exp & Walker2d   & $\mathbf{110.3} \! \pm \! 0.3$ & $\mathbf{110.2} \! \pm \! 0.1$  & $109.9 \! \pm \! 0.3$ \\ 
Medium     & HalfCheetah   & $49.6 \! \pm \! 0.8$ & $\mathbf{50.2} \! \pm \! 0.1$ & $49.9 \! \pm \! 0.5$ \\ 
Medium     & Hopper   & $87.1 \! \pm \! 1.2$ & $\mathbf{96.9} \! \pm \! 1.3$  & $86.8 \! \pm \! 1.4$ \\  
Medium     & Walker2d   & $\mathbf{86.7} \! \pm \! 1.3$ & $\mathbf{86.5} \! \pm \! 0.2$  & $84.9 \! \pm \! 1.3$ \\
Med-Rep & HalfCheetah   & $\mathbf{48.1} \! \pm \! 0.9$ & $47.6 \! \pm \! 0.1$ & $47.2 \! \pm \! 0.7$ \\ 
Med-Rep & Hopper   & $88.9 \! \pm \! 0.8$ & $\mathbf{101.4} \! \pm \! 0.6$  & $83.7 \! \pm \! 2.6$ \\  
Med-Rep & Walker2d   & $87.7 \! \pm \! 1.0$ & $94.0 \! \pm \! 0.2$  & $\mathbf{94.8} \! \pm \! 0.1$ \\ \hline
\multicolumn{2}{c}{\textbf{Average}} & $85.2$         & $\mathbf{88.2}$         & $84.7$ \\
\bottomrule
\end{tabular}
\end{adjustbox}
\end{table}



\section{Related Works}
\textbf{Offline Reinforcement Learning. } 
Offline reinforcement learning (RL) has emerged as a popular approach, allowing agents to learn from pre-existing datasets rather than online interactions with the environment.  However, this method presents challenges, such as optimizing policies using limited demonstrations and mitigating overestimation bias caused by the absence of exploration.  Value-based offline RL approaches tackle these issues through policy regularization techniques, utilizing Q-learning to achieve Bellman optimality from offline trajectories \cite{iql,kumar2020conservative}. These techniques optimize policies to avoid inexperienced actions and evaluate behaviors using learned value functions trained on 'in-sample' data. On the other hand, model-based offline RL approaches adopt a trajectory-planning perspective, utilizing sequence models to generate plausible trajectories \cite{janner2021offline, diffuser,chen2021decisiontransformer}. These model-based approaches have shown promise on tasks with extended horizons. In this paper, we propose a novel hierarchical offline RL approach that combines the strengths of both methods to leverage their respective benefits.

\textbf{Diffusion-based Models for Planning.} Diffusion-based models have excelled in various domains by effectively modeling complex distributions and generating high-quality samples  \cite{ho2020denoising, rombach2021highresolution, song2020denoising}. 
In the context of reinforcement learning, diffusion-based models offer valuable applications for planning by incorporating their ability to model world dynamics \cite{diffuser, decision_diffuser} and also notable expressivity\cite{wang2023diffusion, chi2023diffusion}.  Recent works have extended their model to hierarchical architectures \cite{chen2024simple,wenhao2023hdoffline,hong2023diffused}, showcasing their effectiveness in scalable environments, including long-horizon scenarios. Our proposed model follows a similar approach, but we present an improved model that outperforms existing models in short-horizon and dense-reward setups while remaining competitive in long-horizon and sparse-reward setups.

\textbf{Hierarchical Planning.} Hierarchical planning problems in RL can be divided into two streams: \textit{parallel} and \textit{sequential} planning. 
Sequential planning involves predicting future states sequentially, often utilizing recurrent models or temporal generative models \cite{hmrnn,vta,clockwork_vae,director} 
On the other hand, parallel planning aims to compute all timesteps of future horizons simultaneously. This approach commonly employs generative sequence modeling techniques such as Diffuser \cite{diffuser, decision_diffuser}. 
Recent studies have explored the application of the parallel approach in both subgoal plans and low-level trajectories using hierarchical architecture \cite{hdmi,chen2024simple}. These works have demonstrated that their approach achieves superior performance, especially in long-term tasks. However, our study reveals that these methods fall short in short-horizon and dense-reward setups compared to the performance of recent state-of-the-art flat learning models.
Based on these observations, our work adopts the parallel approach for high-level subgoal plans and incorporates low-level Q-learning-based models to address the limitations in short-horizon and dense-reward setups.

\section{Conclusion and Limitations}
In this work, we present PlanDQ, a novel hierarchical planning method for offline reinforcement learning. By leveraging the strengths of diffusion-based sequence modeling and value learning, PlanDQ demonstrates improved performance across a wide range of tasks, excelling in long-horizon and short-horizon settings. Our analysis highlights the limitations of diffusion-based sequence modeling in noisy, short-horizon tasks, where value learning methods perform more effectively. Our findings provide valuable insights into designing efficient offline RL algorithms, paving the way for further advancements in this crucial field.

\textbf{Limitations and Future Works. }Although PlanDQ presents a promising solution for offline RL, our work has several limitations that should be acknowledged.

First, our approach using a diffusion-based high-level planner results in slower policy generation than value-based methods. Future work should focus on improving the efficiency of PlanDQ without sacrificing performance. 
Second, the quality of the high-level plan in PlanDQ heavily relies on the quality of the data, which could potentially impact the performance. Further investigation into noise-robust high-level planners is necessary to mitigate this limitation. 
Lastly, our setup utilizes the fixed sub-goal interval. Considering adaptive sub-goal intervals could enhance our method to accommodate the varying characteristics of different tasks.

\nocite{langley00}

\section*{Impact Statement}
The research presented in this paper has the potential to make a significant impact on the fields of reinforcement learning and hierarchical planning. From a theoretical perspective, our work offers a deeper understanding of diffusion-based hierarchical planning within the context of offline reinforcement learning, providing new insights by introducing a novel direction that combines diffusion-based sequence modeling with value function learning methods. On a practical level, the PlanDQ method improves the performance of agents in handling complex tasks, potentially contributing to the development of more proficient autonomous systems in various domains, such as healthcare, autonomous driving, and robotics. 

However, with these advancements come potential challenges and responsibilities. As these models become more complex and capable, it is crucial to address the ethical considerations that arise. These include potential biases in decision-making processes, concerns related to data usage and privacy, and the potential displacement of jobs due to increased automation. Ensuring that the ongoing development of offline reinforcement learning is conducted in a responsible and ethical manner is essential as we explore and realize its full potential. 

Moreover, it is essential to ensure that the benefits of our work are accessible to all, with a focus on addressing rather than exacerbating existing inequalities and disadvantages faced by marginalized groups. Collaboration among researchers, policymakers, and industry stakeholders is vital for aligning the development and deployment of these models with societal values and fostering a more equitable and inclusive world. 

\section*{Acknowledgement}
This work is supported by Brain Pool Plus (BP+) Program (No. 2021H1D3A2A03103645) through
the National Research Foundation of Korea (NRF) funded by the Ministry of Science and ICT.

\bibliography{ref_cc, refs_ahn,refs_ahn_local, refs_cg, refs}
\bibliographystyle{icml2024}

\newpage
\appendix
\onecolumn
\section{Benchmark Overview}

\begin{figure}[h]
\begin{center}
\centerline{\includegraphics[width=0.4\linewidth]{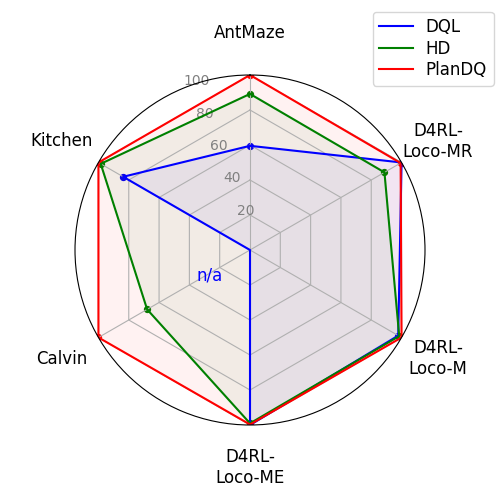}}
\vspace{-1em}
\caption{\textbf{Benchmark Performance. }The graph presents performance results comparing three main baselines: DQL, HD, and our proposed approach, PlanDQ. The evaluation is conducted across six environments: three long-horizon tasks (AntMaze, Kitchen, and Calvin) and three short-horizon tasks (D4RL-Locomotion-Medium-Replay, Medium, and Medium-Expert). Notably, the scores are normalized to the maximum performance of each benchmark.
}
\vspace{-3em}
\label{benchoverview}
\end{center}
\end{figure}

\begin{figure}[h!]
\centering
\begin{minipage}{0.3\linewidth}
     \includegraphics[width=\linewidth]{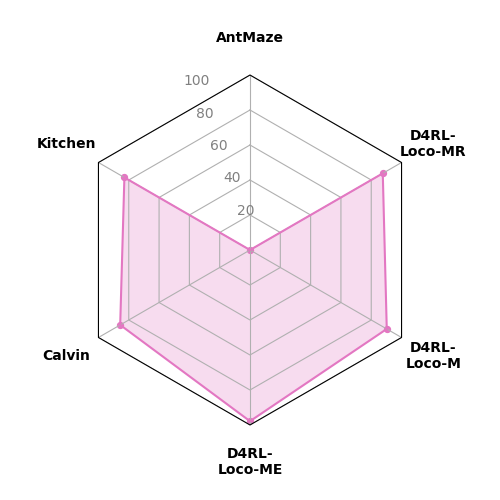}
      \subcaption{DD}
      \label{fig:subfig1}
\end{minipage}
\begin{minipage}{0.3\linewidth}
    \includegraphics[width=\linewidth]{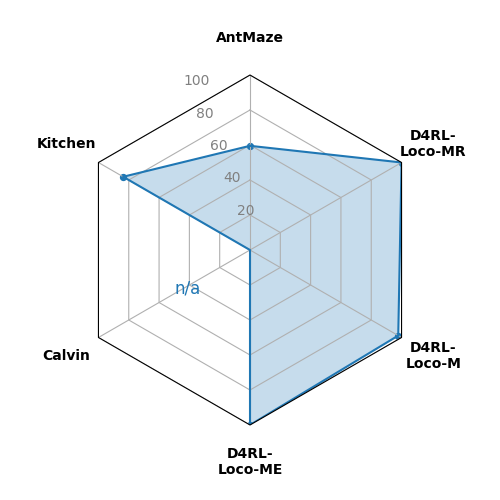}
    \subcaption{DQL}
    \label{fig:subfig2}
\end{minipage}
 \begin{minipage}{0.3\linewidth}
    \includegraphics[width=\linewidth]{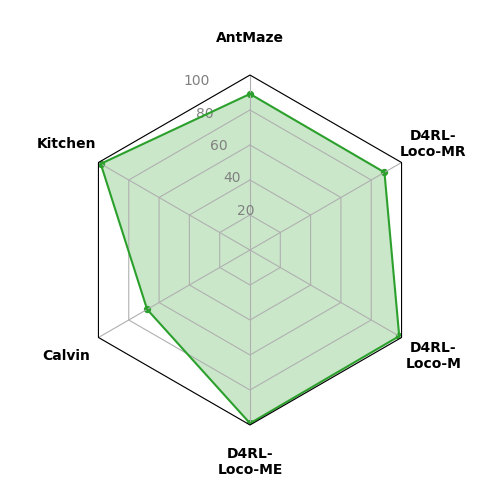}
    \subcaption{HD}
    \label{fig:subfig4}
 \end{minipage}
 \begin{minipage}{0.3\linewidth}
    \includegraphics[width=\linewidth]{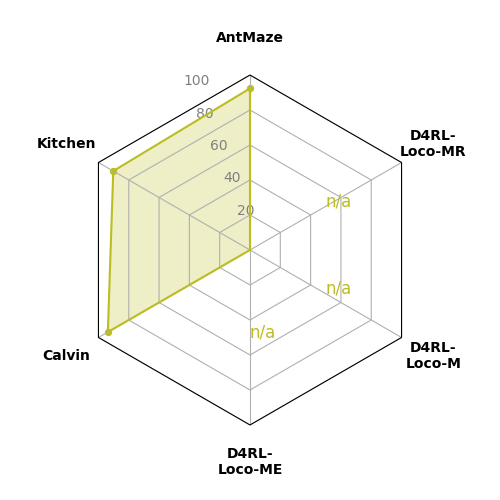}
    \subcaption{HIQL}
    \label{fig:subfig5}
 \end{minipage}
 \begin{minipage}{0.3\linewidth}
    \includegraphics[width=\linewidth]{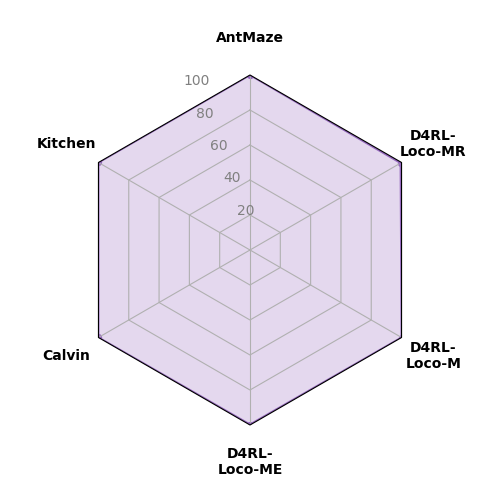}
    \subcaption{PlanDQ(Ours)}
    \label{fig:subfig6}
 \end{minipage}
\caption{Benchmark Overview(Separated Version)}
\end{figure}

\section{An Example-based Analysis for Comparing the Optmality of Diffuser and Q-Learning}\label{app:theory_discuss}
Example \ref{example:1} demonstrates the advantage of the $Q$-learning over the diffuser for short-horizon problems with dense data. In Example \ref{example:1},  the diffuser degrades as the quality of the data-generating policy decreases, whereas 
   $Q$-learning  learns an optimal policy regardless of the quality of the data-generating policy as long as we have dense data: 
\begin{example} \label{example:1}
Consider a MDP, $M=\{S,A,P,R,\gamma,s_{0}\}$ where $s_{0}=c$, $S=\{c\}$ (i.e., $P(c, b,c)=1$ for $b \in A$), $A=\{b_{1},b_{2}\}$, and $R(c,b_{2})>R(c,b_{1})>0$. Then,  the optimal policy $\pi^{*}$ is obtained by $\pi^{*}(b_1|c)=0$ and $\pi^{*}(b_2|c)=1$. Let  $n_{1}$ and $n_2$ be the numbers of observations of taking action $b_1$ and $b_2$ in the training dataset. By chain rule, we can write $p_\theta(\rvx)=\prod_{i} p_\theta(\rvs_{i},\rva_i|\rvs_{1:i-1},\rva_{1:i-1})=\prod_{i} p_\theta(\rva_i|\rvs_{1:i-1},\rva_{1:i-1},\rvs_{i})p_\theta(\rvs_{i}|\rvs_{1:i-1},\rva_{1:i-1})$.
Thus, in this setting, the global optimizer of the training objective of diffuser $p^*_\theta(\rvx)$ is obtained by $p_\theta(b_{i}|\rvs_{1:i-1},\rva_{1:i-1},\rvs_{i})=\frac{n_i}{n_1+n_2}$ for $i \in \{1,2\}$ (and $p_\theta(c|\rvs_{1:i-1},\rva_{1:i-1})=1$). By incorporating an arbitrary guidance function, the policy-induced by the optimal diffuser $\hat \pi$ is $\hat \pi(b_{i}|c)=\frac{n_i}{n_1+n_2}g(c, b_i)$ for $i \in \{1,2\}$, where $g(c, b_i)$ is a factor that corresponds to a chosen guidance function. Thus, $\hat \pi(b_1|c)>\hat \pi(b_2|c)$  if $\frac{n_{1}}{n_{2}}>\frac{g(c, b_2)}{g(c, b_1)}$. This shows that for any $g$ satisfying $\frac{g(c, b_2)}{g(c, b_1)}<\infty$, the policy-induced by the optimal diffuser is arbitrarily far away from the optimal policy if $\frac{n_{1}}{n_{2}}$ is sufficiently large. Here, the ratio $\frac{n_{1}}{n_{2}}$ represents the degree of the suboptimality of the data-generating policy. The condition of $\frac{g(c, b_2)}{g(c, b_1)}<\infty$ is satisfied by common guidance functions (e.g., the value function, the Q-function, the exponential of rewards, etc) since $R(c,b_{1})>0$. On the other hand, the policy induced by the $Q$-learning is  the optimal policy $\pi^{*}$, as long as $n_1,n_2 \ge 1$, regardless of the quality $\frac{n_{1}}{n_{2}}$ of the data generating policy. \qed 
\end{example}
More generally, 
   $Q$-learning is designed to learn a  policy better than a data-generating policy \citep{levine2020offline,fujimoto2021minimalist}. In contrast, the quality of the diffuser is proportional to the quality of the generating policy since the one factor of the diffuser is designed to maximize the likelihood of the given data. This shows that  
   $Q$-learner is better, for example, if we have dense data generated by a poor policy.

When we expand all the value functions, the     
   $Q$-function contains sums over all paths (weighted with the transition probabilities) starting with the current state for the given horizon length.
The number of all such paths increases exponentially as the horizon length increases linearly. Thus, the data tends to be denser for shorter horizons. Thus, on the one hand,   
   $Q$-learner  tends to be advantageous for shorter horizons.

 On  the other hand, the exponential increase in the number of paths  means that learning a sufficiently accurate      
   $Q$-function
requires a significantly larger amount of data for a longer horizon.
However, the diffuser only fits the distribution of data trajectories instead of the entire tree of such paths. \ This can result in a good policy without requiring a significantly larger amount of data for a longer horizon when the data-generating policy is not too poor. 
Thus, diffuser tends to be advantageous for longer horizons.

\section{Implementation Details} \label{app:implementation}
In this section, we describe the details of implementation and hyperparameters we used during our experiments. We have released our code at \url{https://github.com/changchencc/PlanDQ/tree/antmaze}.
\begin{itemize}
    \item We build our D-Conductor upon the officially released Diffuser code obtained from \url{https://github.com/jannerm/diffuser}. 
    \item We build our Q-Performer with the official code obtained from \url{https://github.com/Zhendong-Wang/Diffusion-Policies-for-Offline-RL}.
    \item We set $K = 30$ for the long-horizon planning tasks, while for the Gym-MuJoCo and Kitchen, we use $K = 4$.
    \item The planning horizon is set to $H \cdot K=270$ for AntMaze-Medium, $H \cdot K=450$ for AntMaze-Large, and $H \cdot K=720$ for AntMaze-Ultra. For short-horizon Gym-MuJoCo, we use $H \cdot K=32$. For Calvin, the planning horizon is $H \cdot K=360$. And in Kitchen we use $H \cdot K=64$.
    \item For the MuJoCo locomotion and Kitchen tasks, we select the guidance scales $\omega$ from a set of choices, $\{0.1, 0.01, 0.001, 0.0001\}$, during the planning phase.
    \item In navigation tasks, the intrinsic reward is calculated as the exponential of the negative Euclidean distance between the state $\rvs_{t+1}$, reached after executing action $\rva_t$, and the sub-goal state $\rvs_g$: $r^{\text{intr}}(\rvs_{t}, \rva_t, \rvs_g) = e^{- \lVert \rvs_{t+1} - \rvs_g \rVert ^2}$. We use the cosineMax proposed in Director \citep{director} as intrinsic reward for other tasks.

\end{itemize}

\section{Additional Experiment Results}
\subsection{Ablation on the Low-Level Performer Variants}
We provide additional results comparing PlanDQ with variants where the low-level performer is trained using behavior cloning methods on the AntMaze-Ultra and Kitchen tasks. As shown in Table \ref{table:ablation_bc_addtional}, PlanDQ outperforms its BC variant due to the expressive diffusion-based policy and Q-learning, which utilizes the external task reward for efficient learning.

\begin{table}[h!]
\small
\centering
\caption{\textbf{Ablation Study on Low-Level Performer.} PlanDQ outperforms variants with behavior cloning performers.}
\begin{adjustbox}{width=0.4\linewidth}
\label{table:ablation_bc_addtional}
\begin{tabular}{llccr}
\toprule 
\multicolumn{2}{c}{\textbf{Environment}}         & BC                   & PlanDQ         \\ \hline
AntMaze &Ultra   & $40.0 \! \pm \! 15.4$ & $\mathbf{66.7} \! \pm \! 12.2$ \\ 
Kitchen &Mixed   & $41.7 \! \pm \! 5.1$  & $\mathbf{71.7} \! \pm \! 2.7$ \\  
Kitchen &Partial   & $28.3 \! \pm \! 8.8$  & $\mathbf{75.0} \! \pm \! 7.1$ \\  \hline
\end{tabular}
\end{adjustbox}
\end{table}

\subsection{Ablation on the Jumpy Steps $K$}
We present the results from PlanDQ with different sub-goal time interval $K$ values on the AntMaze tasks in Table \ref{table:ablation_k_ant}. PlanDQ performs well with $K \in {15, 30}$. However, performance drops at $K=45$. This decline may be due to the larger time interval resulting in the loss of key intermediate states, making it challenging for the high-level planner to accurately capture the distribution of sub-goal sequences.
\begin{table}[h!]
\small
\centering
\caption{\textbf{Ablation Study on Sub-goal Time Interval Step $K$.} PlanDQ performs reasonably well with different interval $K$. Results are averaged over 15 planning seeds.}
\begin{adjustbox}{width=0.5\linewidth}
\label{table:ablation_k_ant}
\setlength{\tabcolsep}{4.5pt}
\begin{tabular}{llccr}
\toprule 
\multicolumn{2}{c}{\textbf{Environment}}  & K$15$         & K$30$          & K$45$          \\ \hline
AntMaze & Ultra   &$60.0 \pm 12.6$	&$\textbf{66.7} \pm 12.2$	&$66.7 \pm 12.2$ \\ 
AntMaze & Large   &$80.0 \pm 10.3$	&$\textbf{86.7} \pm 8.7$	    &$60.0 \pm 12.6$ \\  \hline
\multicolumn{2}{c}{\textbf{Average}}   & $70.0$	&$\textbf{76.7}$	&$63.4$ \\ 
\bottomrule
\end{tabular}
\end{adjustbox}
\end{table}

\section{Planning with PlanDQ}
The details of the classifier-guidance sampling process with PlanDQ are presented in Algorithm \ref{app:algo_plandq_planning}.
\begin{algorithm}[h]
\caption{Planning with PlanDQ}
\label{app:algo_plandq_planning}
\begin{algorithmic}[1]
    \STATE {\bfseries Input:} Current State $\rvs$, High-level Diffuser $\bm{\mu}_{\theta}$, guidance function $\mathcal{J}_{\phi}$, guidance scale $\omega$, variance $\sigma_m^2$, low-level policy $\pi_\theta$
    
    \STATE Initialize plan $\rvx^g_M \sim \mathcal{N}(\mathbf{0}, \mathbf{I})$
    
    \WHILE{not done}
        \STATE // \textit{Sample the high-level sub-goal plan}
        \FOR{$m = M-1, \dots, 1$}
        \STATE $\tilde{\bm{\mu}} \leftarrow \bm{\mu}_{\theta} (\rvx^g_{m+1}) + \omega \sigma_m^2\nabla_{\rvx^g_m}\mathcal{J}_{\phi}(\rvx^g_m)$
        \STATE $\rvx^g_{m-1} \sim \mathcal{N}(\tilde{\bm{\mu}}, \sigma_m^2 \mathbf{I})$
        \STATE Fix $\rvg_0$ in $\rvx^g_{m-1}$ to current state $\rvs$
        \ENDFOR
        
        \STATE // \textit{Sample action from low-level policy}
        \STATE Sample $\rva \sim \pi_\theta(\rva | \rvs, \rvg_1)$
        \STATE Execute action $\rva$ in the environment
        \STATE Observe state $\rvs$
    \ENDWHILE
\end{algorithmic}
\end{algorithm}

\section{Sources of Baseline Performance}\label{app_sec:source_baseline}
We detailed the sources for our baseline comparisons below:
\begin{itemize}
    \item \textbf{AntMaze Medium and Large}: We sourced the Diffuser and HD results from \cite{chen2024simple}. The results for IQL, DD, and DTAMP were taken from \cite{hong2023diffused}, and HIQL from \cite{park2023hiql}. Given that we utilized the version 2 (v2) dataset, and to ensure comparability with the original paper which utilized version 0 (v0), we re-ran DQL with the official code.
    \item \textbf{AntMaze Ultra}: We derived the IQL and HIQL results from \cite{park2023hiql}. Diffuser, DD, DQL, and HD were assessed using either their official code or the code provided by the authors.
    \item \textbf{Kitchen}: For Kitchen tasks, DQL results were obtained from \cite{wang2023diffusion}, Diffuser and HD from \cite{chen2024simple}, IQL and DTAMP from \cite{hong2023diffused}, and HIQL from \cite{park2023hiql}. DD results were sourced from \cite{decision_diffuser}.
    \item \textbf{Calvin}: IQL and HIQL results were sourced from \cite{park2023hiql}. We evaluated Diffuser, DD, and HD using either the official code or the code provided by the authors.
    \item \textbf{Gym Tasks}: Results for CQL, IQL, and DQL were taken from \cite{wang2023diffusion}, HD from \cite{chen2024simple}, DTAMP from \cite{hong2023offline}, DD from \cite{decision_diffuser}, TT from \cite{janner2021offline}, and HDMI from \cite{li2022hierarchical}.

\end{itemize}

\end{document}